\documentclass{article}

    \usepackage[preprint]{neurips_2020}

    \usepackage[maxbibnames=99]{biblatex}
    \renewbibmacro{in:}{%
  \ifentrytype{article}{}{\printtext{\bibstring{in}\intitlepunct}}}
    \usepackage{neurips_2020} % I added this (double check)

\usepackage[utf8]{inputenc} % allow utf-8 input
\usepackage[T1]{fontenc}    % use 8-bit T1 fonts
\usepackage{hyperref}       % hyperlinks
\usepackage{url}            % simple URL typesetting
\usepackage{booktabs}       % professional-quality tables
\usepackage{amsfonts}       % blackboard math symbols
\usepackage{nicefrac}       % compact symbols for 1/2, etc.
\usepackage{microtype}      % microtypography

\usepackage{longtable}
\usepackage{graphicx}
\usepackage{xcolor}
\usepackage{algorithm}
\usepackage{algpseudocode}

\title{Combining Domain-Specific Meta-Learners in the Parameter Space for Cross-Domain Few-Shot Classification}

\author{%
  Shuman Peng \\%\thanks{Use footnote for providing further information
  School of Computing Science\\
  Simon Fraser University\\
   \And
   Weilian Song \\
  School of Computing Science \\
  Simon Fraser University \\
   \And
   Martin Ester \\
  School of Computing Science \\
  Simon Fraser University \\
}

\begin{document}

\maketitle

\begin{abstract}
%   The abstract paragraph should be indented \nicefrac{1}{2}~inch (3~picas) on
%   both the left- and right-hand margins. Use 10~point type, with a vertical
%   spacing (leading) of 11~points.  The word \textbf{Abstract} must be centered,
%   bold, and in point size 12. Two line spaces precede the abstract. The abstract
%   must be limited to one paragraph.

The goal of few-shot classification is to learn a model that can classify novel classes using only a few training examples. Despite the promising results shown by existing meta-learning algorithms in solving the few-shot classification problem, there still remains an important challenge: how to generalize to unseen domains while meta-learning on multiple seen domains? In this paper, we propose an optimization-based meta-learning method, called Combining Domain-Specific Meta-Learners (CosML), that addresses the cross-domain few-shot classification problem. CosML first trains a set of meta-learners, one for each training domain, to learn prior knowledge (i.e., meta-parameters) specific to each domain. The domain-specific meta-learners are then combined in the \emph{parameter space}, by taking a weighted average of their meta-parameters, which is used as the initialization parameters of a task network that is quickly adapted to novel few-shot classification tasks in an unseen domain. Our experiments show that CosML outperforms a range of state-of-the-art methods and achieves strong cross-domain generalization ability. 
%In {\color{red} one of the experiments}, CosML even outperforms a baseline method trained for within-domain few-shot classification, demonstrating the effectiveness of leveraging domain-specific knowledge by combining meta-learners in the parameter space.

\end{abstract}

\section{Introduction}

% NeurIPS requires electronic submissions.  The electronic submission site is
% \begin{center}
%   \url{https://cmt3.research.microsoft.com/NeurIPS2020/}
% \end{center}

% Please read the instructions below carefully and follow them faithfully.

Deep neural networks have achieved great success in the supervised learning setting when trained with large amounts of labeled data. However, they lack the ability to generalize to novel tasks when presented with only a small amount of data. This problem setting is commonly known as few-shot learning \cite{lake2015, lake2011oneshot, koch2015siamese, miller2000learningoneexample, triantafillou2017few}. Meta-learning, also known as learning to learn \cite{thrun2012learningtolearn}, addresses this problem by using the meta-learner to produce prior knowledge in order to enable the learner to rapidly adapt to new tasks when presented with only a few labeled examples \cite{vanschoren2018survey}. Three main approaches of meta-learning include metric-based \cite{vinyals2016matchingnet, snell2017protonet, sung2018relationnet}, model-based \cite{santoro16mann, munkhdalai2017metanet}, and optimization-based \cite{ravi2017OptimizationAA, finn2017maml, rusu2018metalearning, nichol2018reptile} frameworks.

Most of the state-of-the-art meta-learning methods \cite{vinyals2016matchingnet, snell2017protonet, finn2017maml, vuorio2019multimodal} rely on the target task to be similar to tasks that have been previously seen during meta-training in order to be able to leverage the prior experiences effectively. In particular, the novel target tasks need to be from the same domain(s) as the training tasks. 
Recently, there has been an emergence of work that explicitly addresses the cross-domain (domain generalization) scenario in few-shot classification \cite{chen2019closerlook, tseng2020crossdomainfewshot, triantafillou2019metadataset, park2019mxml}, where meta-learning models are able to generalize to new datasets. In these studies, the novel task during meta-testing is from some unseen domain which was not used in the meta-training stage. Despite increasing efforts by recent works to improve the domain generalization abilities of few-shot learning, the problem of how to effectively meta-learn across multiple diverse domains without hurting the model’s performance still remains an important challenge \cite{triantafillou2019metadataset}. In this paper, we explore the following research challenges: (1) How can we improve the ability of the meta-learner to meta-learn across multiple heterogeneous domains? (2) How can we generalize the model to unseen domains for optimization-based meta-learning approaches? We will use the terms domain and dataset interchangeably in this paper.

We propose a novel meta-learning method, called \textbf{\underline{C}}ombining D\textbf{\underline{o}}main-\textbf{\underline{S}}pecific \textbf{\underline{M}}eta-\textbf{\underline{L}}earners (CosML), which addresses these challenges, adopting a deep neural network architecture consisting of a feature extractor and a task subnetwork. CosML first trains the domain-specific meta-learners for the seen domains. 
%The domain-specific meta-learners are regularized using mixed-tasks, which simulate novel tasks from an unseen domain, to improve their cross-domain generalization ability. 
When presented with a novel task from an unseen domain, CosML combines the domain-specific meta-learners by taking a weighted combination of the meta-parameters to initialize the task subnetwork, which is then tuned using the small support set of the novel task. Throughout this paper, we refer to an episodically trained model, which in our case is the task subnetwork, as a meta-learner that learns the meta-parameters; the meta-parameters are used to derive the initialization parameters of the task subnetwork for a new task.

CosML adopts the model-agnostic optimization-based meta-learning approach pioneered by MAML \cite{finn2017maml}, while using the notion of prototypes in the feature space from ProtoNet \cite{snell2017protonet} to represent each training domain. 
Additionally, CosML follows the pre-training and meta-training procedure from \cite{chen2019closerlook}. Different from MAML and \cite{chen2019closerlook}, CosML trains a separate meta-learner for each training domain. Furthermore, CosML aggregates the domain-specific meta-learners in the \emph{parameter space}, similar to the idea of the Stochastic Weight Averaging (SWA) procedure proposed by \cite{izmailov2018averaging}. %, which is shown to lead to better generalization performance.

We make the following contributions in this work: (1) We propose an optimization-based meta-learning method, called CosML, that combines meta-learners from seen domains in the \emph{parameter space} to generalize to unseen domains. (2) We introduce mixed tasks to meta-training in order to simulate novel tasks from an unseen domain and to regularize the domain-specific meta-learners. (3) We show strong empirical results for the cross-domain generalization ability of CosML in comparison to the state-of-the-art few-shot classification baselines.

\section{Related work}

Metric-based and optimization-based methods are widely used in recent few-shot learning work, and they can be categorized into two areas: within-domain generalization and cross-domain generalization.

% \paragraph{Within-domain generalization}
Within-domain generalization refers to models that adapt to target tasks that are from the same domain(s) used in the meta-training stage. ProtoNet \cite{snell2017protonet} is a robust metric-based approach that performs nearest neighbour classification on a learned feature space using the Euclidean distance.
MAML \cite{finn2017maml}, LEO \cite{rusu2018metalearning}, and MMAML \cite{vuorio2019multimodal} are optimization-based meta-learning methods that aim to seek an initialization of parameters for a model \(f\) which can be adapted to a novel task with a small number of update steps. 
MMAML is able to effectively meta-learn across multiple domains by modulating the meta-learned prior (initialization) parameters based on the identified task distribution.

% \paragraph{Cross-domain generalization}
Cross-domain generalization refers to models that can effectively adapt to target tasks from an unseen domain, which is a domain that is not used during the meta-training stage. To address cross-domain few-shot learning, Chen et al. \cite{chen2019closerlook} proposes to use a pre-training and fine-tuning training procedure. A feature extractor network \(f_\theta\) is non-episodically trained during pre-training, 
using large amounts of training examples. In the fine-tuning stage, the pre-trained \(f_\theta\) is fixed and only the classifier is trained on few-shot learning tasks.  Tseng et al. \cite{tseng2020crossdomainfewshot} integrates feature-wise transformation layers into the feature extractor of metric-based methods such that diverse feature distributions can be produced during meta-training as a way to capture unseen feature distributions.
In addition to initializing the weights of the feature extractor, Proto-MAML \cite{triantafillou2019metadataset} also initializes the weights of the linear classification layer, which are obtained using ProtoNet. MxML \cite{park2019mxml} consists of an ensemble of meta-learners, where each meta-learner is trained on a different training dataset. In contrast to MxML, our method \emph{combines the parameters} learned by each meta-learner to initialize the model to be finetuned for a novel task rather than combining the predictions made independently by each meta-learner. 
% More concretely, CosML aggregates the domain-specific meta-learners in the parameter space while MxML aggregates the meta-learners in the output space. 

% \paragraph{Averaging model parameters}
Stochastic Weight Averaging (SWA) \cite{izmailov2018averaging} is a deep neural network training procedure which takes the running average of SGD weights during model training by aggregating in the \emph{parameter space} rather than in the output space (i.e., model predictions), leading to better generalization performance. Similar to SWA, we take the weighted average of the parameters across a set of domain-specific neural networks, but the average we take is of the parameters across models from different domains, rather than a running average of parameters proposed by SGD over time for one and the same domain.

%  -- FIGURE 1 -- %
\begin{figure}[]
  \centering
  \includegraphics[scale=0.37]{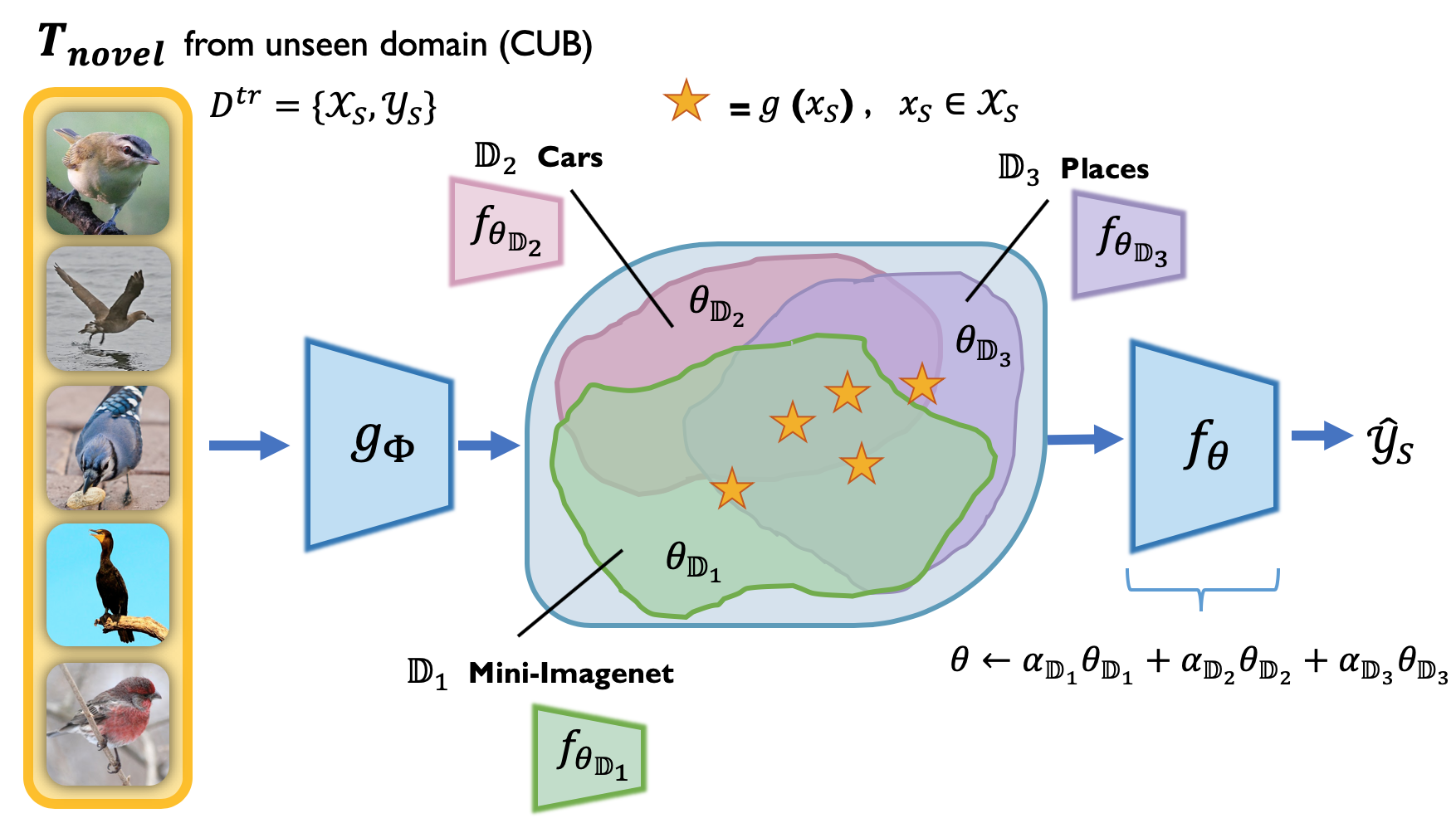}
  \caption{\textbf{Problem definition and method overview.} This figure illustrates CosML performing 5-way 1-shot classification on a novel task \(T_{novel}\) from the unseen domain (dataset) CUB. The model is meta-trained on the seen domains Mini-ImageNet, Cars, and Places. The training (support) set \(D^{tr}\) of \(T_{novel}\) consists of a set of 5 training examples \(\mathcal{X_S}\) (1 example per class) and their corresponding labels \(\mathcal{Y}_S\). The training examples \(x_S \in \mathcal{X}_S\) are used to compute the similarity-based weights \(\alpha_{\mathbb{D}_1}, \alpha_{\mathbb{D}_2}, \alpha_{\mathbb{D}_3}\) for combining the domain-specific meta-parameters \( \theta_{\mathbb{D}_1}, \theta_{\mathbb{D}_2}, \theta_{\mathbb{D}_3}\). \(\theta\) is the initialization parameters for the task subnetwork \(f_\theta\) of \(T_{novel}\), which can quickly adapt to \(T_{novel}\) and predict class labels \(\hat{\mathcal{Y}}_S\). Figure best viewed in color.}
\end{figure}

% \section{Method}

% \subsection{Problem definition}
\section{Problem definition}
\subsection{Background}
We formally define the few-shot learning problem as an \(N\)-way \(K\)-shot classification problem, where each task \(T_i\) is an \(N\)-class classification problem sampled from a task distribution \(p(T)\). Each learning task \(T_i\sim p(T)\) consists of a training (support) set % \(D^{train}=\{S_c =(x_i, y_i) | y_i = c; i=1,...,K\}\),  where \(x_i\) is an example with corresponding class \(y_i\) 
\(D_{T}^{tr}\footnote{We omit the subscript \(i\) from \(T_i\) in \(D_{T_i}^{tr}\) and \(D_{T_i}^{val}\) for clarity.} = \{S_c =(x_j, y_j) | y_j = c; j=1,...,K\}\), where \(x_j\) is an example with corresponding class \(y_j\) and \(K\) is the number of training examples from each class \(c\in\{1,..., N\}\), and a validation (query) set %\(D^{val}=\{S_c= (x_j, y_j) | y_j = c; j=1,..., Q \}\), 
\(D_{T}^{val} = \{S_c= (x_j, y_j) | y_j = c; j=1,..., Q \}\), where \(Q\) is an arbitrary number of validation examples from the same set of \(N\) classes in \(T_i\). \(S_c\) denotes a set of examples with label \(y = c, c\in\{1,...,N\}\).

Following Tseng et al. \cite{tseng2020crossdomainfewshot}, we define a \textbf{seen domain} \(\mathbb{D}^{seen}\) as a domain that is used in the meta-training stage and an \textbf{unseen domain} \(\mathbb{D}^{unseen}\) as a domain that is used exclusively in the meta-testing stage. Moreover, the examples in the unseen domain are not accessible by the model during the meta-training stage.

\subsection{Problem definition}

We define the problem as follows:
Suppose we train a model on \(M\) different seen domains \(\mathbb{D}_1^{seen}, \mathbb{D}_2^{seen}, \dots , \mathbb{D}_M^{seen} \) in the meta-training stage, where each domain has an associated task distribution \(p(T_{\mathbb{D}_k^{seen}}), k=1,...,M\). Let \(\mathbb{D}^{unseen}\) be an unseen domain with task distribution \(p(T_{\mathbb{D}^{unseen}})\). Our goal is to learn a model that can generalize to novel tasks \(T_{novel} \sim p(T_{\mathbb{D}^{unseen}})\) during the meta-testing stage using the meta-learned prior knowledge from each of the \(M\) seen domains. {\bf Figure 1} illustrates this problem setting.

% \begin{figure}[h]
%   \centering
%   \includegraphics[scale=0.4]{figures/figure2_model.png}
%   \caption{\textbf{Model Overview.}  {\color{red}}  }
% \end{figure}

% \subsection{Combining Domain-Specific Meta-Learners (CosML)}
\section{Combining Domain-Specific Meta-Learners (CosML)}

Our goal is to improve the cross-domain generalization ability of the model for few-shot classification on unseen domains. In order to achieve this goal, we propose to exploit the meta-parameters of all the different seen domains as a way to enable the few-shot classification model to generalize to novel target tasks from some unseen domain. In particular, our proposed solution takes a weighted combination of the domain-specific meta-parameters from each of the different seen domains to initialize a model that is then optimized for a novel target task from an unseen domain by performing a few steps of gradient descent.

Our deep neural network model consists of two subnetworks: a feature extractor \(g_\Phi\) that extracts task-invariant features (see section 4.1) and a task subnetwork \(f_\theta\) (see section 4.2) that extracts task-specific features and performs few-shot classification. CosML consists of three stages: pre-training, meta-training, and meta-testing. We explain each stage in detail in the following sections.

\subsection{Pre-training}

We apply a non-episodic approach to first pre-train a neural network to perform image classification, similar to the approach of \cite{chen2019closerlook}. In our experiments, the non-episodic training set consists of the Mini-ImageNet dataset. 
% The non-episodic training set can be any combination of the seen domains, but in our experiments we test two combinations: Mini-ImageNet only and all seen domains. 
After pre-training, we remove the final classification layer as well as the last two hidden layers of the network. This resulting network is the feature extractor \(g_\Phi\), which is then fixed to be used for subsequent stages. Our objective is to learn a non-task-specific representation for examples, in order to extract meaningful feature representations from the examples in a novel target task from some unseen domain \(\mathbb{D}^{unseen}\) during meta-testing. Furthermore, we use the feature space determined by \(g_\Phi\) to compare the similarity between novel tasks from an unseen domain and training tasks from the seen domains, which we will discuss in section 4.2.1.

\subsection{Meta-training}

The aim of meta-training is to train a set of domain-specific meta-learners that learn the meta-parameters, which are combined using similarity-based weights to initialize a task subnetwork \(f_\theta\) for novel tasks from an unseen domain during meta-testing. The weights are determined by the similarity between the novel task and the computed domain and task prototypes belonging to each of the seen domains. The high-level meta-training procedure is shown in \textbf{Algorithm 1}.

% ---- Algorithm 1 ---- % 
\begin{algorithm}
    \caption{CosML meta-training}
    \begin{algorithmic}[1]
        \State\noindent\textbf{Require:} Pure task distributions \(p(T_{\mathbb{D}_1}^p), \dots, p(T_{\mathbb{D}_M}^p)\) and mixed task distribution \(p(T^m)\)
        \State\textbf{Require:} Hyper-parameters \(\gamma, \beta\), pre-trained feature extractor \(g_\Phi\)
        \State\noindent\textbf{Output:} \(M\) %domain-specific 
        meta-parameters \(\theta_{\mathbb{D}_k} \) and prototypes \(P_{\mathbb{D}_k} = \{Z_{T\mathbb{D}_k}, Z_{\mathbb{D}_k}\}\) for \(k=1,\dots,M\)
        \State Randomly initialize \(\theta_{\mathbb{D}_1}, \dots, \theta_{\mathbb{D}_M} \)
        \While{not done}
            % \State// Set 1: pure tasks //
            \For{each seen domain \(\mathbb{D}_k , k = 1,\dots,M\)} 
            \hspace{8em} \(\triangleright\) Set 1: pure tasks \(\triangleleft\)
            \State Sample a batch of pure tasks \(T_j^p \sim p(T^p_{\mathbb{D}_k})\)
            \For{each task \(T_j^{p} = (D_{T_j^p}^{tr}, D_{T_j^p}^{val})\)}
                \State Compute new task prototype \(z_{T_j}\) using equation 1 and append it to \(Z_{T\mathbb{D}_k}\)
                \State Update \(Z_{\mathbb{D}_k}\) using equation 2
                % \State Evaluate \( \nabla_{\theta_{\mathbb{D}_k}}\mathcal{L}_{T_j^p}(f_{\theta_{\mathbb{D}_k}}(g_\Phi(x)); D_{T_j^p}^{tr})\) w.r.t. \(K\) examples per class
                \State Compute adapted parameters with gradient descent using K examples per class: \[ \theta'_{T_j^p} = \theta_{\mathbb{D}_k} - \gamma\nabla_{\theta_{\mathbb{D}_k}}\mathcal{L}_{T_j^p}(f_{\theta_{\mathbb{D}_k}}(g_\Phi(x)); D_{T_j^p}^{tr})\]
            \EndFor
            \State Update \(\theta_{\mathbb{D}_k} \leftarrow \theta_{\mathbb{D}_k} - \beta\nabla_{\theta_{\mathbb{D}_k}}\sum_{T_j^p \sim p(T_{\mathbb{D}_k}^p)} \mathcal{L}_{T_j^p}(f_{\theta'_{T_j^p}}(g_\Phi(x)); D_{T_j^p}^{val})\)
            \EndFor

            % \State // Set 2: mixed tasks //
            \State Sample a batch of mixed tasks \(T_j^m \sim p(T^m)\) 
            \hspace{8em} \(\triangleright\) Set 2: mixed tasks \(\triangleleft\)
            
            \For{each task \(T_j^m = (D_{T_j^m}^{tr}, D_{T_j^m}^{val})\)}
                \State Compute \(\alpha_{\mathbb{D}_k{T_j^m}}\) for \(k=1,\dots,M\) using equations 5 and 6
                \State Compute \(\theta = \alpha_{\mathbb{D}_1{T_j^m}}\theta_{\mathbb{D}_1} + \dots + \alpha_{\mathbb{D}_M{T_j^m}}\theta_{\mathbb{D}_M} \) 
                % \State Evaluate \( \nabla_{\theta}\mathcal{L}_{T_j^m}(f_{\theta}(g_\Phi(x)); D_{T_j^m}^{tr})\) w.r.t. \(K\) examples per class
                \State Compute adapted parameters with gradient descent using K examples per class: \[ \theta'_{T_j^m} = \theta - \gamma\nabla_{\theta}\mathcal{L}_{T_j^m}(f_{\theta}(g_\Phi(x)); D_{T_j^m}^{tr}) \]
            \EndFor 
            % \State\noindent end for
            \State Update each \(\{\theta_{\mathbb{D}_k} | k = 1,\dots,M\}\) using equation 7
            
            % \[ \theta_{\mathbb{D}_k} \leftarrow \theta_{\mathbb{D}_k} - \beta\nabla_{\theta_{\mathbb{D}_k}} \sum_{T_j^m \sim p(T^m)} \alpha_{\mathbb{D}_k T_j^m}\mathcal{L}_{T_j^m}(f_{\theta'_{T_j^m}}(g_\Phi(x)); D_{T_j^m}^{val})\]
        \EndWhile
    \end{algorithmic}
\end{algorithm}

\subsubsection{Prototypes}
Similar to and following the notations from \cite{snell2017protonet}, we use \emph{task prototypes} to represent training tasks used during meta-training in the feature space. Furthermore, we use \emph{domain prototypes} to represent different domains in the feature space.

Let us first denote \(S_{T\mathbb{D}_k}= \{T_i \;|\; T_i = \{D_{T_i}^{tr}, D_{T_i}^{val}\}, i=1,2, ... \}\) as a set of training tasks from domain \(\mathbb{D}_k\) used during meta-training; \(S_{T\mathbb{D}_k}\) grows with the number of training iterations. Let \(S_\mathbb{D} =\{(x, y) \in D_{T_i}^{tr} \cup\; D_{T_i}^{val} \;|\; T_i \in S_{T\mathbb{D}_k} \}\)  be the set of all the training task examples belonging to domain \(\mathbb{D}_k\). Finally, let \(P_{\mathbb{D}_k} = \{Z_{T\mathbb{D}_k}, Z_{\mathbb{D}_k} \}\)  denote the set of prototypes for domain \(\mathbb{D}_k\), which includes a growing set of task prototypes \(Z_{T\mathbb{D}_k}\) and one domain prototype \(Z_{\mathbb{D}_k}\) that is updated during meta-training. %Note that we compute one task prototype for each training task and one domain prototype for each seen domain, using the examples in the training tasks.
We formally define a \emph{task prototype} \(z_T \in Z_{T\mathbb{D}} \) as the average feature vector of all of the training (support)  and validation (query) examples in a given training task \(T\) that belongs to a specific domain \(\mathbb{D}^{seen}\): 
% \[ z_T = \frac{1}{|D_T^{tr}\cup D_T^{val}|} \sum_{x \;\in D_T^{tr} \cup D_T^{val}} g_\Phi(x)\]
\begin{equation}
    z_T = \frac{1}{|D_T^{tr}\cup D_T^{val}|} \sum_{x \;\in D_T^{tr} \cup D_T^{val}} g_\Phi(x)
\end{equation}

We also formally define a \emph{domain prototype} as the average feature vector of all the examples from the training and validation set of training tasks used during meta-training from a given domain \(\mathbb{D}^{seen}\):\footnote{We omit the superscript \(seen\) from \(\mathbb{D}^{seen}\) for clarity.}
% \[Z_\mathbb{D} = \frac{1}{|S_\mathbb{D}|} \sum_{x \;\in S_\mathbb{D}} g_\Phi(x) \]
\begin{equation}
    Z_\mathbb{D} = \frac{1}{|S_\mathbb{D}|} \sum_{x \;\in S_\mathbb{D}} g_\Phi(x)
\end{equation}

% A task prototype is specific to some domain and it represents examples belonging to a specific training task of a given domain. Although the purpose of task and domain prototypes may seem redundant -- ultimately they both abstractly represent examples from the same domain -- it is still necessary to explicitly use task prototypes because different training tasks in the same domain can have very different prototypes. 

% \paragraph{Computing prototypes adaptively}
Task and domain prototypes are \textbf{adaptively computed} during meta-training as new training tasks are sampled and used to train the model. The set of task prototypes grows with an increasing number of training tasks, while the number of domain prototypes remains constant, although the domain prototypes themselves change. %There is only one domain prototype per domain and there are \(N\) task prototypes for each domain, where \(N\) is the number of training tasks (episodes) {\color{red}from each domain} used in the meta-training stage. 
% The intuition of adaptively computing task and domain prototypes during the meta-training stage is that the training tasks that are used to train the model during the meta-training stage have a direct influence on the meta-learned parameters, and therefore they should be represented in the feature space when computing the similarities between tasks and domains. 

\paragraph{Similarity between tasks and prototypes}
We define the distance between a task  \(T\) and a prototype as the average distance between the training examples in the task and the prototype in the feature space determined by \(g_\Phi\):% R^D \rightarrow R^M\):
\begin{equation}
    ds(T, z) = \frac{1}{|D_T^{tr}|} \sum_{x_i \;\in D_T^{tr}} d(g_\Phi(x_i), z)
\end{equation}

where \(z\) % \in R^M\) 
is a (task or domain) prototype and \(D_T^{tr}\) is the training set of task \(T\). \(T\) is a mixed task \(T^m\) (see section 4.2.2) during the meta-training stage and a novel task \(T_{novel}\) at meta-test time. Finally, \(d(\cdot, \cdot)\) is a distance function such as the Euclidean distance, which we are using in our implementation. 

To determine the distance between a task and a seen domain \(\mathbb{D}_k^{seen}\), we consider both the distance to the domain prototype and the distance to all task prototypes corresponding to \(\mathbb{D}_k^{seen}\)\footnote{We again omit the superscript \(seen\) from \(\mathbb{D}_k^{seen}\) for clarity.} as follows:
\begin{equation}
    dist(T, \mathbb{D}_k) = \frac{1}{2} \times [ds(T, Z_{\mathbb{D}_k}) + \frac{1}{N_{Tk}} \sum_{z_T \in Z_{T\mathbb{D}_k}} ds(T, z_T)] 
\end{equation}

where \(N_{Tk}\) is the total number of training tasks used so far in \(\mathbb{D}_k^{seen}\). We compute \(dist(T, \mathbb{D}_k)\) for all the seen domains, \(k=1,...,M\). 

The domain weight \(\alpha_{\mathbb{D}_k}\) assigned to a seen domain \(\mathbb{D}_k^{seen}\) is inversely proportional to its distance; the closer the task is to a domain in the feature space, the larger the weight will be for that domain: 
   \begin{equation}
       \alpha_{\mathbb{D}_k} = 1/dist(T, \mathbb{D}_k)
   \end{equation}

To combine the domain-specific meta-learners (task subnetworks) by taking a weighted average of their meta-parameters, we normalize the domain weights as follows:
    \begin{equation}
        \alpha_{\mathbb{D}_k} = \alpha_{\mathbb{D}_k} / \sum_{j=1}^{M} \alpha_{\mathbb{D}_j}
    \end{equation}

% \paragraph{Pure tasks and mixed tasks}
\subsubsection{Pure tasks and mixed tasks}
Following the common practice of meta-learning, we use \textbf{pure tasks} -- tasks that consist of classes (and examples) from a single domain -- to episodically train a set of model parameters that can be quickly adapted to different tasks from the same domain. In addition, we propose to use \textbf{mixed tasks} -- tasks that consist of randomly selected classes from different domains -- in order to simulate novel tasks from an unseen domain and to update and regularize the domain-specific meta-parameters so that they will be able to adapt better to unseen domains in the meta-test phase.  

In each iteration of our episodic training, we consider two sets of tasks. The first set consists of a mini-batch of pure tasks \(T^p\), where each task includes a training set \(D_{T^p}^{tr}\) and a corresponding validation set \(D_{T^p}^{val}\). We first use the set of pure tasks to learn domain-specific meta-parameters, using the same meta-training procedure as MAML. The second set consists of a mini-batch of mixed tasks \(T^m\), where each task also includes \(D_{T^m}^{tr}\) and \(D_{T^m}^{val}\). We treat the set of mixed tasks as novel tasks from some unseen domain and initialize their task subnetwork \(f_\theta\) using the weighted average of the domain-specific meta-parameters \(\theta_{\mathbb{D}_1}, \dots , \theta_{\mathbb{D}_M}\) as follows: 
% using the weighted average of the meta-parameters \(\theta_{\mathbb{D}_1}, \dots , \theta_{\mathbb{D}_M}\) of the  existing domain-specific task subnetworks as follows: 
\( \theta = \alpha_{\mathbb{D}_1}\theta_{\mathbb{D}_1} + \dots + \alpha_{\mathbb{D}_M}\theta_{\mathbb{D}_M}\).
    % \begin{equation} 
    %     \theta = \alpha_{\mathbb{D}_1}\theta_{\mathbb{D}_1} + \dots + \alpha_{\mathbb{D}_M}\theta_{\mathbb{D}_M}
    % \end{equation}

Based on the performance of \(f\) on the mixed tasks after a few gradient steps using \(D_{T^m}^{tr}\),  each of the domain-specific meta-parameters \( \theta_{\mathbb{D}_k} \)will be updated accordingly: 
\begin{equation}
    %  \theta_{\mathbb{D}_k} \leftarrow \theta_{\mathbb{D}_k} - \beta\nabla_{\theta_{\mathbb{D}_k}} \sum_{T_j^{m}} \alpha_{\mathbb{D}_{kT_j^m}} L_{T_j^m}(f(x;\theta_{T_j^m}'; D_{T_j^m}^{val})
    \theta_{\mathbb{D}_k} \leftarrow \theta_{\mathbb{D}_k} - \beta\nabla_{\theta_{\mathbb{D}_k}} \sum_{T_j^m \sim p(T^m)} \alpha_{\mathbb{D}_k T_j^m}\mathcal{L}_{T_j^m}(f_{\theta'_{T_j^m}}(g_\Phi(x)); D_{T_j^m}^{val})
\end{equation}
where \(\beta\) is the same meta step size used by MAML. 
In the above update step, the loss is computed by evaluating the adapted model \(f\), using updated task-specific parameters \(\theta'_{T_j^m}\), on the validation set of mixed task \(T_j^m\).  The task-specific parameters \(\theta'_{T_j^m}\) are learned using the training set of task \(T_j^m\).

% \subsubsection{Meta-testing}
\subsection{Meta-testing}
At meta-test time, we employ the meta-parameters of the domain-specific task subnetworks obtained from meta-training to learn a task subnetwork for novel tasks from an unseen domain \(\mathbb{D}^{unseen}\). To do so, we initialize a task subnetwork with the weighted average of the domain-specific meta-parameters, weighted by the similarity of these domains and a novel task \(T_{novel}\) from \(\mathbb{D}^{unseen}\). This model is then optimized via a few steps of gradient descent using the training (support) set of \(T_{novel}\). The performance of the final model is evaluated on the test (query) set of \(T_{novel}\).

\section{Experiments}
\subsection{Experimental setup}

\paragraph{Dataset}
We use the Mini-ImageNet \cite{ravi2017OptimizationAA}, CUB \cite{welinder2010cub}, Cars \cite{krause2013cars}, Places \cite{zhou2017places}, and Plantae \cite{vanhorn2018plantae} datasets to evaluate the cross-domain few-shot classification performance of CosML in comparison with existing few-shot learning methods. 
Additional details can be found in the Supplementary Material.
% Details about the datasets can be found in the Supplementary Material. %number of classes in each dataset as well as the train, validation, and test splits used can be found in the Supplementary Material. 

% -- TABLE 1: MAIN RESULTS -- %
\begin{table}[h!]
    \centering
    % \begin{tabular}{c|c} 
    %      &  \\
    %      & 
    % \end{tabular}
    \caption{\textbf{Cross-domain few-shot classification accuracy.} The dataset listed in each column represents the unseen domain %during the meta-testing stage,
    while the remaining 4 datasets are seen domains used during meta-training. % the seen domains used in the meta-training stage. 
    All the methods use the PT-miniImagenet pre-trained feature extractor. % For brevity, we refer to PT-miniImagenet as PT-mini. 
    We include experimental results for MAML initialized with PT-miniImagenet and for MAML without any initialization, denoted by \emph{no-PT}. All the methods use the Conv-4 backbone except for MatchingNet LFT* and RelationNet LFT*, which use the ResNet-10 \cite{he2016resnet} backbone. Results for MatchingNet LFT* and RelationNet LFT* are from \cite{tseng2020crossdomainfewshot}.} 
    % \label{tab:my_label}
    \begin{tabular}{@{}lcccccc@{}} \toprule
    % \begin{tabular}{p{0.4\textwidth}CGC} \toprule
    \multicolumn{5}{c}{5-way 1-shot} \\  \midrule
    % \multicolumn{1}{}{} & \multicolumn{4}{c}{unseen domains}
    % \\ \cmidrule(r){2-5}
    Method & CUB & Cars & Places & Plantae \\ \midrule
    MatchingNet LFT* & \(43.29 \pm 0.59\%\)  & \(30.62 \pm 0.48 \%\) & \(52.51 \pm 0.67\%\) & \(\textbf{35.12} \pm \textbf{0.54}\%\) \\
    RelationNet LFT* & \(\textbf{48.38} \pm \textbf{0.63}\%\) & \(32.21 \pm 0.51\%\) & \(50.74 \pm 0.66\%\) & \(35.00 \pm 0.52\%\) \\
    MatchingNet LFT  \cite{tseng2020crossdomainfewshot} & \( 34.20 \pm 0.53\% \) & \( 30.15 \pm 0.46\% \) & \( 39.43 \pm 0.60\% \)& \(29.50 \pm 0.39\%\) \\
    RelationNet LFT \cite{tseng2020crossdomainfewshot} & \( 39.70 \pm 0.58\% \) & \( 32.59 \pm 0.54\% \) & \( 39.92 \pm 0.59\% \) & \( 33.11 \pm 0.56\% \) \\
    ProtoNet \cite{snell2017protonet} & \( 36.54 \pm 0.52\% \) & \( 29.38 \pm 0.42\% \) & \( 40.12 \pm 0.59\% \) & \( 31.42 \pm 0.49\% \)  \\
    Proto-MAML \cite{triantafillou2019metadataset} & \( 36.05 \pm 0.53\% \) & \( 29.46 \pm 0.44\% \) & \( 38.71 \pm 0.57\% \) & \( 31.20 \pm 0.49\% \) \\
    MAML \cite{finn2017maml} (no PT) & \(35.06 \pm 0.54\%\) & \(31.12 \pm 0.54\%\) & \(36.14\pm0.56\%\) & \(30.95 \pm 0.49\%\)  \\
    MAML \cite{finn2017maml} & \(35.50 \pm 0.53\%\) & \(26.76 \pm 0.42\%\) & \(39.21 \pm 0.60\%\) & \(31.35 \pm 0.49\%\) \\
    \midrule
    \textbf{Ours:} CosML & \(46.89 \pm 0.59\%\) & \(\textbf{47.74} \pm \textbf{0.59}\%\) & \(\textbf{53.96} \pm \textbf{0.62}\%\) & \(30.93 \pm 0.46\%\) \\
    \bottomrule
    \toprule
    \multicolumn{5}{c}{5-way 5-shot} \\
    \midrule
    Method & CUB & Cars & Places & Plantae \\ \midrule
    MatchingNet LFT* & \(61.41 \pm 0.57\%\) & \(43.08 \pm 0.55\%\) & \(64.99 \pm 0.59\%\) & \(48.32 \pm 0.57\%\) \\
    RelationNet LFT* & \(64.99 \pm 0.54\%\) & \(43.44 \pm 0.59\%\) & \(67.35 \pm 0.54\%\) & \(\textbf{50.39} \pm \textbf{0.52}\%\) \\
    MatchingNet LFT \cite{tseng2020crossdomainfewshot} & \(49.09 \pm 0.53\%\) & \(42.42 \pm 0.53\%\) & \(54.15 \pm 0.54\%\) & \(43.32 \pm 0.53\%\) \\
    RelationNet LFT \cite{tseng2020crossdomainfewshot} & \(55.53 \pm 0.59\%\) & \(46.05 \pm 0.55\%\) & \(53.17 \pm 0.55\%\) & \(45.66 \pm 0.56\%\) \\
    ProtoNet \cite{snell2017protonet} & \(56.37 \pm 0.53\%\) & \(43.83 \pm 0.55\%\) & \(59.91 \pm 0.56\%\) & \(\textbf{50.39} \pm \textbf{0.59}\%\) \\
    Proto-MAML \cite{snell2017protonet} & \(57.21 \pm 0.54\%\) & \(45.06 \pm 0.56\%\) & \(58.38 \pm 0.57\%\) & \(47.45 \pm 0.55\%\) \\
    MAML \cite{finn2017maml} (no PT) & \(53.20 \pm 0.54\%\) & \(43.71 \pm 0.56\%\) & \(53.91 \pm 0.57 \%\) & \(44.70 \pm 0.53\%\) \\
    MAML \cite{finn2017maml} & \(52.66 \pm 0.52\%\) & \(43.43 \pm 0.53\%\) & \(56.61 \pm 0.58\%\) & \(42.72 \pm 0.55\%\) \\
    \midrule
    \textbf{Ours:} CosML & \(\textbf{66.15} \pm \textbf{0.63}\%\) & \(\textbf{60.17} \pm \textbf{0.63}\%\) & \({\color{black}\textbf{88.08} \pm \textbf{0.46\textbf{}}}\%\) & \(42.96 \pm 0.57\%\) \\
    
    \bottomrule    
    \end{tabular}
\end{table}

% % -- TABLE 2: MAML VS COSML -- %
% \begin{table}[h!]
%     \centering
%     % \begin{tabular}{c|c} 
%     %      &  \\
%     %      & 
%     % \end{tabular}
%     \caption{\textbf{Comparing CosML’s cross-domain few-shot classification performance with the within-domain few-shot classification performance of MAML.} MAML is trained and tested on the same dataset, whereas CosML is trained on all datasets except the unseen test dataset, which appears in each column. %A total of 40,000 episodes are used to train each of the models. 
%     5-way 5-shot classification performance is shown.}
%     % \label{tab:my_label}
%     \begin{tabular}{@{}lcccccc@{}} \toprule
%     Method  & CUB & Cars & Places & Plantae \\ \midrule
%     MAML & \(\textbf{74.38} \pm \textbf{0.52}\%\) & \(\textbf{62.39} \pm \textbf{0.57}\%\) & \(64.53 \pm 0.61\%\) & \(\textbf{63.61} \pm \textbf{0.60}\%\) \\
%     CosML (PT-mini) & \(66.15 \pm 0.63\%\) & \(60.17 \pm 0.63\%\) & \(\textbf{93.75} \pm \textbf{0.35}\%\) & \(42.96 \pm 0.57\%\) \\
%     \bottomrule
%     \end{tabular}
% \end{table}

\paragraph{Implementation details}
We use the 4-module convolutional network (Conv-4) architecture that is commonly used in few-shot classification \cite{vinyals2016matchingnet, finn2017maml, snell2017protonet}. 
In our implementation of CosML, the feature extractor \(g_\Phi\) consists of the first two modules, which are fixed, and the task subnetwork \(f_\theta\) consists of the last two modules and a linear classification layer. Additional implementation details can be found in the Supplementary Material.

% The feature space determined by \(g_\Phi\) is {\color{red}28224-dimensional}. 
% All the images are resized to size \(84 \times 84\). 
% Furthermore, the mini-batch sizes {\color{red}(mentioned in section 4.2.2)} that we use for pure tasks and for mixed tasks are \(25\times M\) and \(25\), respectively, where \(M\) is the number of seen domains. A mini-batch of pure tasks contains the same number (25) of tasks from each domain.

\paragraph{Pre-trained feature extractor}
We denote the Conv-4 network that is pre-trained on the Mini-ImageNet dataset as \emph{PT-miniImagenet}. 
% We denote the Conv-4 network that is pre-trained on the Mini-ImageNet dataset as \emph{PT-miniImagenet}. We also evaluate CosML using our own custom pre-trained Conv-4 network that is non-episodically trained using all datasets but the unseen dataset. We refer to our custom pre-trained feature extractor as \emph{PT-LOO} (leave-one-out). % The reason for this additional comparison is due to the fact that our method freezes the layers in the feature extractor in both the meta-training and meta-testing stages, which means that it is important for the feature extractor to be able to extract domain-invariant and task-invariant features prior to the meta-training stage. 
% Since our ultimate goal is to evaluate the model’s few-shot classification performance on an unseen dataset, we can pre-train using the same datasets as those used in the meta-training stage as long as examples from the unseen domain are not accessed. 

\paragraph{Baseline methods}
The selected baseline methods include both within-domain and cross-domain metric- and optimization-based meta-learning methods: MatchingNet and RelationNet with learning-to-learned feature-wise transformation (LFT)\footnote{Implementation from https://github.com/hytseng0509/CrossDomainFewShot.} \cite{tseng2020crossdomainfewshot}, ProtoNet\footnote{Implementation from https://github.com/wyharveychen/CloserLookFewShot. \label{closerlookcode}} \cite{snell2017protonet}, MAML\textsuperscript{\ref{closerlookcode}} \cite{finn2017maml}, and Proto-MAML\footnote{Implementation from https://github.com/google-research/meta-dataset.} \cite{triantafillou2019metadataset}. We initialize each baseline method as well as CosML with PT-miniImagenet. The entire pre-trained Conv-4 network is fine-tuned in the baseline methods, whereas we only use the first two modules of the pre-trained network as the feature extractor for our method and keep them fixed. 

We follow the same leave-one-out experimental setup as \cite{tseng2020crossdomainfewshot}. This means that only 4 out of the 5 datasets are used as seen domains during meta-training; the held out domain becomes the unseen domain. 
We do not tune any of the hyper-parameters for our method. 
For the baseline methods, we use the hyper-parameters that are provided in their original implementations. More experimental details can be found in the Supplementary Material.

% -- ABLATION TABLE -- %
\begin{table}[]
    \centering
    \caption{\textbf{Ablation studies.} The accuracies shown are for leave-one-out 5-way 5-shot cross-domain few-shot classification. CosML\(_1\) uses uniform weights instead of similarity-based weights. CosML\(_2\) is meta-trained without mixed tasks. CosML\(_3\) uses a deeper feature extractor \(g_\Phi\); \(g_\Phi\) contains 3 conv modules and \(f_\theta\) consists of 1 conv module and a classification layer. The PT-miniImagenet pre-trained feature extractor is used.}
    % \label{tab:my_label}
    \begin{tabular}{@{}lcccccc@{}} \toprule
    Method  & CUB & Cars & Places & Plantae \\ \midrule
    CosML\(_1\) (uniform \(\alpha_k\))  & \(86.46 \pm 0.47\%\) & \(57.71 \pm 0.60\%\) & \(72.36 \pm 0.58\%\) & \(44.37 \pm 0.55\%\) \\
    CosML\(_2\) (no \(T^m\))  & \(35.26 \pm 0.43\%\) & \(32.24 \pm 0.41\%\) & \(31.95 \pm 0.42\%\) & \(26.27 \pm 0.35\%\) \\
    CosML\(_3\) (deeper \(g_\Phi\)) & \(52.21 \pm 0.59\%\) & \(41.21 \pm 0.50\%\) & \(52.10 \pm 0.61\%\) & \(26.51 \pm 0.36\%\) \\
    Complete CosML & \(66.15 \pm 0.63\%\) & \(60.17 \pm 0.63\%\) & \({\color{black}88.08 \pm 0.46}\%\) & \(42.96 \pm 0.57\%\) \\
    \bottomrule
    \end{tabular}
\end{table}

\subsection{Experimental results}

\paragraph{Main results}
Table 1 presents the accuracy of CosML and all the baseline methods for 5-way 1-shot and for 5-way 5-shot classification.  All of the reported results represent the mean accuracy with a \(95\%\) confidence interval of 1000 randomly sampled novel tasks from the selected unseen domain. We observe that CosML consistently outperforms all baseline methods for all unseen domains except for Plantae. {\color{black}We hypothesize the lack of performance improvement for CosML on the unseen domain Plantae is due to a large domain difference between the Plantae dataset and the other datasets. Metric-based baselines -- MatchingNet LFT*, RelationNet LFT*, and ProtoNet -- perform the best on the unseen domain Plantae. This is consistent with the experimental results from \cite{chen2019closerlook, triantafillou2019metadataset}, where metric-based methods are shown to outperform optimization-based methods when the domain difference is large. 
The performance of CosML is consistent in both the 5-way 1-shot and 5-shot settings. 
{\color{black}Our experimental results demonstrate the effectiveness of leveraging domain-specific knowledge by combining meta-learners in the parameter space. }

\paragraph{Ablation studies}
The results of our ablation studies are reported in Table 2. 
The performance decreases significantly without the use of mixed-tasks during meta-training. On the CUB and Plantae datasets, we observe better performance when using uniform weights instead of similarity-based weights to combine the domain-specific meta-learners. While this confirms the effectiveness of combining meta-learners, this also suggests that better similarity-based weighting schemes should be investigated in future research.
Finally, we observe that increasing the depth of \(g_\Phi\) by 1 conv module and decreasing the depth of \(f_\theta\) by 1 conv module for CosML hurts the cross-domain performance substantially.
{\color{black} This observation aligns with the experimental findings from \cite{yosinki2014how}. Features that are more specific to the Mini-ImageNet (pre-training) dataset are transferred when we increase the depth of our pre-trained and fixed feature extractor $g_\Phi$, which hurts the performance when we train the remaining layers of the network on a different dataset. }

\section{Conclusion}
In this paper, we proposed a novel method for few-shot classification in the cross-domain setting, called Combining Domain-Specific Meta-Learners (CosML), that leverages the meta-learned knowledge from each of the seen training domains. More specifically, CosML combines the meta-learners by taking a weighted average of the domain-specific meta-parameters, which is used to initialize a new task subnetwork to quickly adapt to a novel task from an unseen domain. We show strong empirical results for CosML in comparison to the state-of-the-art within-domain and cross-domain few-shot learning methods. This shows the effectiveness of leveraging domain-specific knowledge by combining meta-learners in the parameter space.

To the best of our knowledge, CosML is the first meta-learning method that combines model parameters to support quick adaptation to tasks from an unseen domain. Not only is our method simple, it is also effective in a variety of settings. 
% We believe that the combination of meta-learned prior knowledge in the parameter space has the potential to make meta-learning and few-shot learning applicable in more realistic, diverse, and challenging scenarios.
For future work, we believe it is important to investigate how to best divide the neural network into the feature extractor subnetwork and task subnetwork, i.e., how many fixed layers to use for the feature extractor and how many layers to use and train for the task subnetwork. Also, as CosML does not know the confidence of its cross-domain predictions, a method needs to be developed to assess the similarities of the novel tasks from an unseen domain to the seen domain(s) in order to compute confidences, which we leave for future research.

\newpage
\section*{Acknowledgement}
We would like to thank Shao-Hua Sun, Hossein Sharifi-Noghabi, Xiang Xu, Jialin Lu, and Yudong Luo for the discussions and support. We would also like to thank Compute Canada for providing the computational resources. This research was supported by the Natural Sciences and Engineering Research Council (NSERC) Discovery Grant "Data mining in heterogeneous information networks with attributes" (to Martin Ester).

% \section*{References}

% References follow the acknowledgments. Use unnumbered first-level heading for
% the references. Any choice of citation style is acceptable as long as you are
% consistent. It is permissible to reduce the font size to \verb+small+ (9 point)
% when listing the references.
% {\bf Note that the Reference section does not count towards the eight pages of content that are allowed.}
\medskip

\small

% [1] Alexander, J.A.\ \& Mozer, M.C.\ (1995) Template-based algorithms for
% connectionist rule extraction. In G.\ Tesauro, D.S.\ Touretzky and T.K.\ Leen
% (eds.), {\it Advances in Neural Information Processing Systems 7},
% pp.\ 609--616. Cambridge, MA: MIT Press.

% [2] Bower, J.M.\ \& Beeman, D.\ (1995) {\it The Book of GENESIS: Exploring
%   Realistic Neural Models with the GEneral NEural SImulation System.}  New York:
% TELOS/Springer--Verlag.

% [3] Hasselmo, M.E., Schnell, E.\ \& Barkai, E.\ (1995) Dynamics of learning and
% recall at excitatory recurrent synapses and cholinergic modulation in rat
% hippocampal region CA3. {\it Journal of Neuroscience} {\bf 15}(7):5249-5262.

% \bibliography{cosml_neurips_2020.bib}
\printbibliography

@inproceedings{yosinki2014how,
  title={How transferable are features in deep neural networks?},
  author={Jason Yosinki and Jeff Clune and Yoshua Bengio and Hod Lipton},
  booktitle={Advances in Neural Information Processing Systems},
  year={2014}
}

@inproceedings{deng2009imagenet,
  title={ImageNet: A large-scale hierarchical image database},
  author={Jia Deng and Wei Dong and Richard Socher and Li-Jia Li and Kai Li and Li Fei-Fei},
  booktitle={IEEE Conference on Computer Vision and Pattern Recognition},
  year={2009}
}

@article{hilliard2018cubsplits,
    title={Few-shot learning with metric-agnostic conditional embeddings},
    author={Nathan Hilliard and Lawrence Phillips and Scott Howland and Artem Yankov and Courtney D Corley and Nathan O Hodas},
    year={2018},
    journal={arXiv preprint arXiv:1802.04376}
}

@article{kingma2014adam,
  title={Adam: A method for stochastic optimization},
  author={Kingma, Diederik P and Ba, Jimmy},
  journal={arXiv preprint arXiv:1412.6980},
  year={2014}
}

@article{lake2015,
	author = {Brenden M Lake and Ruslan Salakhutdinov and Joshua B Tenenbaum},
	title = {Human-level concept learning through probabilistic program induction},
	volume = {350},
	number = {6266},
	pages = {1332--1338},
	year = {2015},
	doi = {10.1126/science.aab3050},
	publisher = {American Association for the Advancement of Science},
% 	issn = {0036-8075},
% 	URL = {https://science.sciencemag.org/content/350/6266/1332},
% 	eprint = {https://science.sciencemag.org/content/350/6266/1332.full.pdf},
	journal = {Science}
}

@article{vanschoren2018survey,
    title={Meta-Learning: A Survey},
    author={Joaquin Vanschoren},
    year={2018},
    journal={arXiv preprint arXiv:1810.03548}
}

@incollection{vinyals2016matchingnet,
    title = {Matching Networks for One Shot Learning},
    author = {Vinyals, Oriol and Blundell, Charles and Lillicrap, Timothy and kavukcuoglu, koray and Wierstra, Daan},
    booktitle = {Advances in Neural Information Processing Systems},
    year = {2016}
}

@incollection{snell2017protonet,
    title = {Prototypical Networks for Few-shot Learning},
    author = {Snell, Jake and Swersky, Kevin and Zemel, Richard},
    booktitle = {Advances in Neural Information Processing Systems},
    year = {2017},
}

@INPROCEEDINGS{sung2018relationnet,  
    author={Flood Sung and Yongxin Yang and Li Zhang and Tao Xiang and Philip H.S. Torr and Timothy M. Hospedales},  
    booktitle={IEEE Conference on Computer Vision and Pattern Recognition},   
    title={Learning to Compare: Relation Network for Few-Shot Learning},   
    year={2018}
}

@InProceedings{santoro16mann,
    title = 	 {Meta-Learning with Memory-Augmented Neural Networks},
    author = 	 {Adam Santoro and Sergey Bartunov and Matthew Botvinick and Daan Wierstra and Timothy Lillicrap},
    booktitle = 	 {Proceedings of The 33rd International Conference on Machine Learning},
    year = 	 {2016}
}

@inproceedings{ravi2017OptimizationAA,
    title={Optimization as a Model for Few-Shot Learning},
    author={Sachin Ravi and Hugo Larochelle},
    booktitle={International Conference on Learning Representations},
    year={2017}
}

@InProceedings{finn2017maml,
    title = 	 {Model-Agnostic Meta-Learning for Fast Adaptation of Deep Networks},
    author = 	 {Chelsea Finn and Pieter Abbeel and Sergey Levine},
    booktitle = 	 {International Conference on Machine Learning},
    year = 	 {2017}
 }

@inproceedings{vuorio2019multimodal,
    title={Multimodal Model-Agnostic Meta-Learning via Task-Aware Modulation},
    author={Vuorio, Risto and Sun, Shao-Hua and Hu, Hexiang and Lim, Joseph J.},
    booktitle={Neural Information Processing Systems},
    year={2019},
}

@inproceedings{triantafillou2019metadataset,
    title={Meta-Dataset: A Dataset of Datasets for Learning to Learn from Few Examples},
    author={Eleni Triantafillou and Tyler Zhu and Vincent Dumoulin and Pascal Lamblin and Utku Evci and Kelvin Xu and Ross Goroshin and Carles Gelada and Kevin Swersky and Pierre-Antoine Manzagol and Hugo Larochelle},
    booktitle={International Conference on Learning Representations},
    year={2020},
    % url={https://openreview.net/forum?id=rkgAGAVKPr}
}

@inproceedings{chen2019closerlook,
    title={A Closer Look at Few-shot Classification},
    author={Wei-Yu Chen and Yen-Cheng Liu and Zsolt Kira and Yu-Chiang Frank Wang and Jia-Bin Huang},
    booktitle={International Conference on Learning Representations},
    year={2019},
}

@inproceedings{tseng2020crossdomainfewshot,
    author = {Tseng, Hung-Yu and Lee, Hsin-Ying and Huang, Jia-Bin and Yang, Ming-Hsuan},
    booktitle = {International Conference on Learning Representations},
    title = {Cross-Domain Few-Shot Classification via Learned Feature-Wise Transformation},
    year = {2020}
}

@article{park2019mxml,
    title={MxML: Mixture of Meta-Learners for Few-Shot Classification},
    author={Minseop Park and Jungtaek Kim and Saehoon Kim and Yanbin Liu and Seungjin Choi},
    journal={arXiv preprint arXiv:1904.05658},  
    year={2019}
}

@article{izmailov2018averaging,
    title={Averaging Weights Leads to Wider Optima and Better Generalization},
    author={Izmailov, Pavel and Podoprikhin, Dmitrii and Garipov, Timur and Vetrov, Dmitry and Wilson, Andrew Gordon},
    journal={arXiv preprint arXiv:1803.05407},
    year={2018}
}

@inproceedings{rusu2018metalearning,
    title={Meta-Learning with Latent Embedding Optimization},
    author={Andrei A. Rusu and Dushyant Rao and Jakub Sygnowski and Oriol Vinyals and Razvan Pascanu and Simon Osindero and Raia Hadsell},
    booktitle={International Conference on Learning Representations},
    year={2019},
    % url={https://openreview.net/forum?id=BJgklhAcK7},
}

@article{zhou2017places,
    title={Places: A 10 million Image Database for Scene Recognition},
    author={Zhou, Bolei and Lapedriza, Agata and Khosla, Aditya and Oliva, Aude and Torralba, Antonio},
    journal={IEEE Transactions on Pattern Analysis and Machine Intelligence},
    year={2017},
    publisher={IEEE}
 }

@INPROCEEDINGS{krause2013cars,  
    author={Jonathan Krause and Michael Stark and Jia Deng and Li Fei-Fei},  
    booktitle={IEEE International Conference on Computer Vision Workshops},   
    title={3D Object Representations for Fine-Grained Categorization},   year={2013}, 
}

@techreport{welinder2010cub,
	Author = {Peter Welinder and Steve Branson and Takeshi Mita and Catherine Wah and Florian Schroff and Serge Belongie and Pietro Perona},
	Institution = {California Institute of Technology},
	Number = {CNS-TR-2010-001},
	Title = {{Caltech-UCSD Birds 200}},
	Year = {2010}
}

@inproceedings{vanhorn2018plantae,
    title=  {The inaturalist species classification and detection dataset},
    author={Grant Van Horn and Oisin Mac Aodha and Yang Song and Yin Cui, Chen Sun and Alex Shepard and Hartwig Adam and Pietro Perona and Serge Belongie},
    booktitle=  {IEEE Conference on Computer Vision and Pattern Recognition},
    year={2018}
}

@INPROCEEDINGS{he2016resnet,
  author={Kaiming He and Xiangyu Zhang and Shaoqing Ren and Jian Sun},
  booktitle={IEEE Conference on Computer Vision and Pattern Recognition}, 
  title={Deep Residual Learning for Image Recognition}, 
  year={2016}
}

@article{ioffe2015batch,
    title={Batch Normalization: Accelerating Deep Network Training by Reducing Internal Covariate Shift},
    author={Sergey Ioffe and Christian Szegedy},
    journal={arXiv preprint arXiv:1502.03167},    
    year={2015}
}

@inproceedings{
    munkhdalai2017metanet,
    author={Tsendsuren Munkhdalai and Hong Yu},
    title = {Meta Networks},
    booktitle = {International Conference on Machine Learning},
    year = {2017}
}

@article{
    nichol2018reptile,
    author = {Alex Nichol and John Schulman},
    title = {Reptile: a Scalable Metalearning Algorithm},
    journal ={arXiv preprint arXiv:1803.02999},
    year = {2018}
}

@inproceedings{
    miller2000learningoneexample,
    author={Erik G Miller and Nicholas E Matsakis and Paul A Viola},
    title = {Learning from one example through shared densities on transforms},
    booktitle = {IEEE Conference on Computer Vision and Pattern Recognition},
    year = {2000}
}

@inproceedings{
    lake2011oneshot,
    author={Brenden M. Lake and Ruslan Salakhutdinov and Jason Gross and Joshua B. Tenenbaum},
    title = {One shot learning of simple visual concepts},
    booktitle = {CogSci},
    year = {2011}
}

@techreport{
    koch2015siamese,
    author={Gregory Koch},
    title ={Siamese neural networks for one-shot image recognition},
    number = {Master's thesis},
    institution = {University of Toronto},
    year = {2015}
}

@article{
    thrun2012learningtolearn,
    author = {Sebastian Thrun and Lorien Pratt},
    title = {Learning to learn},
    journal ={Springer Science \& Business Media},
    year = {2012}
}

@inproceedings{triantafillou2017few,
  title={Few-shot learning through an information retrieval lens},
  author={Triantafillou, Eleni and Zemel, Richard and Urtasun, Raquel},
  booktitle={Advances in Neural Information Processing Systems},
  pages={2255--2265},
  year={2017}
}

% ========================== % 
%  APPENDIX                  %
% ========================== % 

\newpage
\appendix

\section{Datasets}

{\color{black}For our cross-domain few-shot classification experiments, we use the Mini-ImageNet, CUB (birds), Cars, Places, and Plantae datasets. We follow the data pre-processing procedure from \cite{ravi2017OptimizationAA} and \cite{hilliard2018cubsplits} for the Mini-ImageNet and CUB datasets respectively. We follow the data pre-processing procedure from \cite{tseng2020crossdomainfewshot} for the Cars, Places, and Plantae datasets. {\bf Table 1} shows a summary of each dataset, which includes the dataset source, the data splits used, as well as the number of classes in each of the train, validation, and test splits.

The Mini-ImageNet dataset contains images of a variety of ojects. The CUB, Cars, Places, and Plantae datasets are fine-grained datasets that contain images of different species of birds, cars, places, and plants respectively. }

\begin{table}[h]
    \centering
    % \begin{tabular}{c|c} 
    %      &  \\
    %      & 
    % \end{tabular}
    \caption{{\color{black}\textbf{Dataset details.} This table shows the dataset source, origin of the dataset splits used in the experiments, as well as the number of classes in each of the train, validation, and test splits. }}
    % \label{tab:my_label}
    \resizebox{\textwidth}{!}{\begin{tabular}{@{}lcccccc@{}} \toprule
    Dataset  & Source & Split used & Train classes & Validation classes  & Test classes \\ \midrule
    Mini-ImageNet & Deng et al. \cite{deng2009imagenet} & Ravi \& Larochelle \cite{ravi2017OptimizationAA} & 64 & 16 & 20 \\
    CUB (birds)  & Welinder et al. \cite{welinder2010cub}  & Hilliard et al. \cite{hilliard2018cubsplits} & 100 & 50 & 50 \\
    Cars & Krause et al. \cite{krause2013cars} & Tseng et al. \cite{tseng2020crossdomainfewshot} & 98 & 49 & 49 \\ 
    Places & Zhou et al. \cite{zhou2017places} & Tseng et al. \cite{tseng2020crossdomainfewshot} & 183 & 91 & 91 \\
    Plantae & Van Horn et al. \cite{vanhorn2018plantae} & Tseng et al. \cite{tseng2020crossdomainfewshot} & 100 & 50 & 50 \\
    \bottomrule
    \end{tabular}}
\end{table}

\section{Additional experimental details}

\subsection{Implementation details}
% The feature space determined by \(g_\Phi\) is {\color{red}28224-dimensional}. 
% All the images are resized to size \(84 \times 84\). 
% Furthermore, the mini-batch sizes {\color{red}(mentioned in section 4.2.2)} that we use for pure tasks and for mixed tasks are \(25\times M\) and \(25\), respectively, where \(M\) is the number of seen domains. A mini-batch of pure tasks contains the same number (25) of tasks from each domain.
 \paragraph{Network architecture}
 {\color{black}We use the 4-module convolutional network (Conv-4) architecture that is commonly used in few-shot classification \cite{vinyals2016matchingnet, finn2017maml, snell2017protonet}. Each of the four modules consists of a \(3\times3\) convolutional layer with 64 output channels, followed by a batch normalization layer \cite{ioffe2015batch}, a ReLU activation function, and finally a \(2\times2\) max pooling layer. Outputs from the last module are 1600-dimensional feature vectors, which are inputs to the linear layer for \(N\)-way classification. In our experiments, we use the same Conv-4 backbone for CosML and for all the baseline methods. 
 
 In our implementation of CosML, the feature extractor \(g_\Phi\) consists of the first two modules and the task subnetwork \(f_\theta\) consists of the last two modules and a linear classification layer. Note that the feature space determined by \(g_\Phi\) is 28224-dimensional. All the images are resized to size \(84 \times 84\). 
 
 We will make the code for the implementation of our proposed method, CosML, publicly available on GitHub.}

\subsection{Hyper-parameters}

{\color{black}\paragraph{Optimizer} In all of our experiments, we use the Adam optimizer \cite{kingma2014adam} with default settings from PyTorch. The hyper-parameter values in the default setting are: the learning rate is 0.001, the betas coefficients are (0.9, 0.999), the eps term is 1e-8, weight\_decay is 0, and the amsgrad flag is set to False.}

\paragraph{Metric-based methods}
{\color{black}We use the same hyper-parameters as Tseng et al. \cite{tseng2020crossdomainfewshot} for training the MatchingNet LFT and RelationNet LFT models on the Conv-4 backbone. We use the same hyper-parameters as Snell et al. \cite{snell2017protonet} for training the ProtoNet models. However, rather than using a higher number of \emph{way} for training the ProtoNet models, we use the same number of \emph{way} (i.e., 5-way) as the tasks we use to evaluate the model during meta-testing to ensure fairness and consistency with the other models. The validation (query) set contains 16 images.}

{\color{black}\paragraph{Optimization-based methods}
We use the same hyper-parameters for MAML, Proto-MAML, and CosML in both the 5-way 1-shot and 5-way 5-shot settings. All the models are trained using a slow outer-loop learning rate (meta-step size) of 0.001, a fast inner-loop learning rate (step size) of 0.01, 5 gradient steps, a meta-batch size of 4 tasks, and a validation (query) set of 16 images. 
For CosML, the mini-batch sizes that we use for pure tasks and for mixed tasks are \(25\times M\) and \(25\), respectively, where \(M\) is the number of seen domains. A mini-batch of pure tasks contains the same number of tasks from each domain.

}

\subsection{Training configurations}
\paragraph{Hardware}
{\color{black}We train all of the models, except for Proto-MAML, on a single NVIDIA V100SXM2 GPU with 16G of memory. Due to insufficient GPU memory, we train all Proto-MAML models on a CPU node with 60G of memory.}

\paragraph{Training}
{\color{black}All the few-shot classification models are trained using a total of 40,000 training tasks. }

\end{document}